\documentclass[conference]{IEEEtran}
\ifCLASSINFOpdf
\else
\fi
\usepackage{amsmath, amssymb, graphicx, xcolor, adjustbox} 

\newcommand{\comma}{\,,}
\newcommand{\point}{\,.}
\usepackage{xspace}
\newcommand{\fps}{FPS\xspace}

\newcommand{\mysection}[1]{\vspace{2pt}\noindent\textbf{#1}}
\usepackage{url}
\usepackage{interval}
\usepackage{hyperref}
\begin{document}
%
\title{Investigating Event-Based Cameras for\\Video Frame Interpolation in Sports}

\author{\IEEEauthorblockN{Antoine Deckyvere}
\IEEEauthorblockA{Montefiore Institute, University of Li\`ege \\ Email: antoine.dekyvere@student.uliege.be}
\and
\IEEEauthorblockN{Anthony Cioppa}
\IEEEauthorblockA{Montefiore Institute, University of Li\`ege\\ Email: anthony.cioppa@uliege.be}
\and ~~~~~~~~~~ \and
\IEEEauthorblockN{Silvio Giancola}
\IEEEauthorblockA{IVUL, KAUST\\Email: silvio.giancola@kaust.edu.sa}
\and
\IEEEauthorblockN{Bernard Ghanem}
\IEEEauthorblockA{IVUL, KAUST\\Email: Bernard.Ghanem@kaust.edu.sa}
\and
\IEEEauthorblockN{Marc Van Droogenbroeck}
\IEEEauthorblockA{Montefiore Institute, University of Li\`ege\\Email: m.vandroogenbroeck@uliege.be}}

%


\maketitle

\begin{abstract}
Slow-motion replays provide a thrilling perspective on pivotal moments within sports games, offering a fresh and captivating visual experience.
However, capturing slow-motion footage typically demands high-tech, expensive cameras and infrastructures. 
Deep learning Video Frame Interpolation (VFI) techniques have emerged as a promising avenue, capable of generating high-speed footage from regular camera feeds. 
Moreover, the utilization of event-based cameras has recently gathered attention as they provide valuable motion information between frames, further enhancing the VFI performances.
In this work, we present a first investigation of event-based VFI models for generating sports slow-motion videos.
Particularly, we design and implement a bi-camera recording setup, including an RGB and an event-based camera to capture sports videos, to temporally align and spatially register both cameras.
Our experimental validation demonstrates that \emph{TimeLens}, an off-the-shelf event-based VFI model, can effectively generate slow-motion footage for sports videos. 
This first investigation underscores the practical utility of event-based cameras in producing sports slow-motion content and lays the groundwork for future research endeavors in this domain.

\end{abstract}

\mysection{Keywords.} Event-based camera, video frame interpolation, video understanding, sports analysis, slow motion.

\section{Introduction}
\label{sec:introduction}

Slow-motion footage has become an indispensable tool in sports broadcasting, serving to highlight or replay pivotal moments with enhanced emotions and accurate details. 
Beyond its aesthetic appeal, slow-motion technology offers a multifaceted utility in sports analysis. By slowing down the action, it provides analysts and coaches with a finer temporal resolution to scrutinize player movements, dissecting techniques and strategies with precision. 
Moreover, slow-motion footage empowers referees to make more informed decisions regarding fouls, as it allows for a comprehensive review of contentious incidents at lower speeds. 
This dual function of slow-motion technology, therefore, both enriches the viewing experience for fans and enhances the decisions of coaches and officials.

\begin{figure}
    \centering
    \includegraphics[width=\columnwidth]{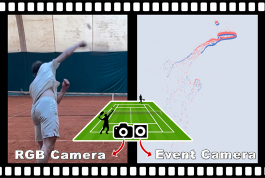}
    \caption{\textbf{Bi-camera recording setup for event-based video frame interpolation in sports.} We propose a two camera recording setup, including an RGB camera and an event-based camera to capture sports videos, and temporally align and spatially register both cameras. We record video footage of racquet sports characterized by high-speed movements of the ball and racquets, utilizing our specialized setup, and demonstrate the effectiveness of off-the-shelf event-based video frame interpolation techniques in producing slow-motion footage.}
    \label{fig:graphical_abstract}
\end{figure}

However, the widespread adoption of slow-motion technology in sports broadcasting has been held back by the prohibitive costs associated with high-speed camera setups, which are typically only feasible for major competitions. 
As a result, lower leagues and smaller-scale sporting events often lack access to this technology, further increasing the accessibility gap in sports media production.
Additionally, the infrastructure required to support high-speed camera systems entails considerable investment, presenting a significant barrier for many organizations.

Fortunately, recent advancements in deep learning started to revolutionize the landscape of slow-motion production. 
New techniques enable Video Frame Interpolation (VFI), aiming to generate intermediate frames produced by standard cameras, circumventing the need for specialized, high-cost equipment. 
Moreover, the emergence of event-based cameras represents a groundbreaking development in this field. 
Unlike conventional cameras that capture entire frames at fixed intervals, event-based cameras detect and record individual pixel-level changes in luminance asynchronously, only transmitting data when a \emph{visual event} occurs. 
This innovative approach reduces data redundancy and provides crucial information on motion dynamics between frames at a much higher frame rate. 
By capturing motion in a continuous fashion, event-based cameras offer unparalleled precision in frame interpolation, enabling the generation of seamless slow-motion sequences with enhanced realism. 

In this study, we investigate the use of an off-the-shelf event-based video frame interpolation method, namely \emph{TimeLens}~\cite{Tulyakov2021TimeLens}, for the creation of slow-motion footage in the sports domain. 
In particular, we design and implement a bi-camera recording setup, including an RGB camera and an event-based camera to capture sports videos. 
We then temporally align and spatially register both cameras to project the information provided by the event-based camera to the RGB frames.
As illustrated in Figure~\ref{fig:graphical_abstract}, our experimental approach entails recording sports video footage, with a particular focus on racquet sports, characterized by high-speed movements of the ball and racquets, utilizing our specialized setup.
Through our analysis, we demonstrate the effectiveness of TimeLens in generating slow-motion sequences, providing valuable insights into the dynamics of sports events.
This study serves as an initial step towards exploring the potential of event-based camera systems in sports video production, laying the groundwork for future research aimed at refining and expanding this technology for broader applications in the sports industry.

\section{Related Work}
\label{sec:related_work}
Our work relates to the research fields of video frame interpolation, event-based cameras, and sports video understanding.

\subsection{Video frame interpolation}

Video Frame Interpolation (VFI) involves generating intermediate frames between existing video frames, effectively increasing the frame rate and allowing the creation of slow-motion content~\cite{Parihar2021AComprehensive}.
Over the years, various methods have been proposed, leveraging convolutional neural networks (CNNs)~\cite{Niklaus2017Video}, phase-based techniques~\cite{Meyer2015Phasebased}, optical flow algorithms~\cite{Niklaus2018ContextAware}, or Generative Adversarial Networks (GANs)~\cite{vanAmersfoort2017Frame-arxiv}.
Moreover, industrial tools such as \emph{XtraMotion} from EVS Broadcast Equipment~\cite{EVS2024XtraMotion} facilitate slow-motion video production in professional settings.
The field can count on the availability of numerous datasets and benchmarks to develop methods, including popular resources such as Vimeo90K~\cite{Xue2019Video}, UCF101~\cite{Soomro2012UCF101-arxiv}, X4K1000FPS~\cite{Sim2021XVFI}, Middlebury~\cite{Baker2007ADatabase}, and the SNU-FILM~\cite{Choi2020Channel} datasets.
Recently, SportsSlowMo~\cite{Chen2023SportsSloMo-arxiv} introduced the first dataset dedicated to human-centric sports video clips for VFI using only the RGB modality. Our study also aims to investigate sports-centric videos, incorporating event-based camera feeds as an additional modality for VFI.

\subsection{Event-based camera}
Event-based cameras, also known as neuromorphic sensors, operate differently from traditional cameras. Rather than capturing entire frames at fixed intervals, event-based cameras only record changes in brightness, known as ``events'', at pixel level. These cameras provide several advantages over conventional cameras, such as a high temporal resolution of up to $10{,}000$ frames per second, low power consumption, wide dynamic range, sparse data output for efficient data transmission and storage, and low latency~\cite{Zheng2023Deep-arxiv}.
Recently, event-based cameras have emerged as integral components in video frame interpolation pipelines thanks to their ability to capture information between RGB frames. This feature enables the modeling of more intricate scenarios, including non-linear motion often present in sports content. Notable methods like Timelens~\cite{Tulyakov2021TimeLens}, Timelens++~\cite{Tulyakov2022TimeLens++}, or SuperFast~\cite{Gao2023SuperFast} have demonstrated impressive performance in generating slow-motion footage from both modalities. However, their application has predominantly focused on close-up shots within controlled environments.
In this study, we investigate the ability of event-based VFI methods to handle sports videos captured in natural, uncontrolled settings.

\subsection{Sports Video understanding}
Over the past decade, there has been an increased focus on sports video understanding research~\cite{Moeslund2014Computer, Naik2022AComprehensive} thanks to the availability of large-scale datasets~\cite{Ingwersen2023SportsPose, Scott2022SoccerTrack, VanZandycke2022DeepSportradarv1, Istasse2023DeepSportradarv2}. 
The SoccerNet datasets and challenges~\cite{Giancola2022SoccerNet, Cioppa2023SoccerNetChallenge-arxiv} have contributed to multiple video understanding tasks in sports such as, action spotting~\cite{Giancola2018SoccerNet}, replay grounding~\cite{Deliege2021SoccerNetv2}, camera calibration and player re-identification~\cite{Cioppa2022Scaling}, multiple player tracking~\cite{Cioppa2022SoccerNetTracking}, multi-view video recognition~\cite{Held2023VARS}, and dense video captioning~\cite{Mkhallati2023SoccerNetCaption}.
Recently, deep learning enabled real-time data and analysis on player performance~\cite{Vandeghen2022SemiSupervised, Seweryn2024Improving-arxiv, Boeker2023Soccer}, including, segmentation and tracking of players and the ball~\cite{Cioppa2019ARTHuS, Maglo2022Efficient}, game tactics~\cite{ArbuesSanguesa2020Using}, action spotting~\cite{Giancola2018SoccerNet, Deliege2021SoccerNetv2,Cioppa2018ABottomUp,Soares2022Temporally,Hong2022Spotting-arxiv, Wu2020Fusing-arxiv, Denize2023COMEDIAN-arxiv, Seweryn2023Survey-arxiv, Cioppa2021Camera}, and the creation of video highlights or summaries ~\cite{Cioppa2020AContextaware, Gautam2022Assisting-RG}. 
Overall, these developments elevate coaching strategies and enhance the overall viewer experience~\cite{Midoglu2022MMSys, Sarkhoosh2024AIBased}. In this work, we show the capabilities of event-based VFI methods to generate high-speed footage for downstream sports video analysis.

\section{Methodology}
\label{sec:method}

In this section, we provide a rigorous definition of the event-based video frame interpolation task, describe our sports data recording setup, and explain the temporal alignment and spatial registration between our bi-camera setup.

\subsection{Event-based video frame interpolation}
\label{subsec:interpolation}

Let $\mathbf{I}(t_i)$ represent the video image recorded at time $t_i$ by a regular RGB camera, where $\mathbf{I}(t_i) \in \mathbb{R}^{H \times W \times C}$ denotes an image frame with height $H$, width $W$, and $C$ channels.
Then, let $\mathbf{E}(t_i)$ be the event data recorded by the event-based camera between time $t_i$ and $t_{i+1}$, consisting of a set of events $\{\mathbf{e}_j\}$, where each event $\mathbf{e}_j = (x_j, y_j, t'_j, p_j)$ corresponds to a change in luminance at pixel location $(x_j, y_j)$ and at time $t'_j$ with polarity $p_j \in \{-1, 1\}$.  

The event-based video frame interpolation problem can be formulated as follows.
Given a sequence of event data $\{\mathbf{E}(t_1), \mathbf{E}(t_2), ..., \mathbf{E}(t_n)\}$, where $t_1 < t_2 < ... < t_n$, and the corresponding image frames $\{\mathbf{I}(t_1), \mathbf{I}(t_2), ..., \mathbf{I}(t_n)\}$, the objective is to estimate some intermediate frame $\hat{\mathbf{I}}(t_{i+\Delta t})$ at any arbitrary time $t_{i+\Delta t} \in \interval[open]{t_{i}}{t_{i+1}} $. Hence, event-based video frame interpolation aims to generate intermediate frames between consecutive frames captured by a regular RGB camera, using the information provided by an event-based camera. 

In this work, we focus on a particular case of event-based video frame interpolation where only the bounding image frames $\mathbf{I}(t_i)$ and $\mathbf{I}(t_{i+1})$, along with the \emph{registered} event data, $\mathbf{E}(t_i)$ are used to compute intermediate frames $\hat{\mathbf{I}}(t_{i+\Delta t})$. 

\subsection{Bi-camera sports data recording setup}

\begin{figure}[t]
\centering
    \includegraphics[width=\columnwidth]{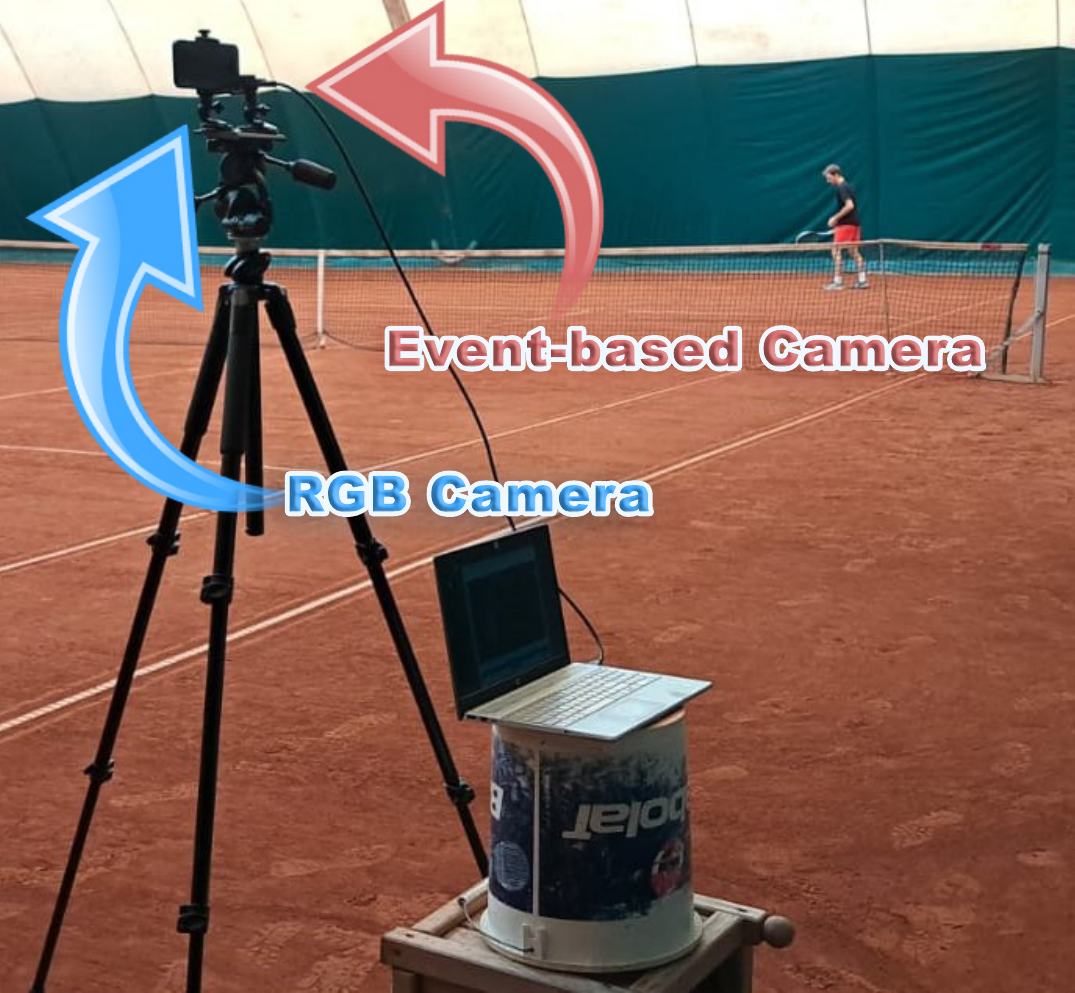}
    \caption{\textbf{Camera setup.} Picture of our bi-camera setup used for sports data collection. We align both the RGB camera and event-based camera on a tripod.}
    \label{fig:camera_setup}
\end{figure}

We design and implement a bi-camera (RGB and event-based) recording setup for sports data collection by aligning two camera supports and mounting them on tripods, as illustrated in Figure \ref{fig:camera_setup}. To ensure accurate alignment, we position the event-based camera such that its optical center is aligned with the center of the RGB camera lens.

For the RGB camera, we use an iPhone $12$, capable of capturing videos at $120$ frames per second (\fps) with a resolution of $1920\times1080$ pixels. For the event-based camera, we employ the Prophesee EVK4, which offers a resolution of $720\times1280$ pixels, a time resolution equivalent to over $10{,}000$ \fps, and a dynamic range exceeding $120$ decibels (dB). Moreover, the event-based camera exhibits a low light cutoff of $0.08$ lux, ensuring reliable performance even in dimly lit conditions. We employed the Metavision software suite for both recording and post-processing of the event-based camera footage directly on-site.

We recorded footage of two semi-professional tennis players engaging in a game on an indoor court illuminated by LED panels to avoid flickering caused by traditional neon lighting. The players were instructed to play a regular game while we moved the cameras between each game session. In total, we recorded approximately $30$ minutes of usable footage captured from the RGB and event-based cameras.

\subsection{Temporal alignment and spatial registration}
\label{subsec:alignment}

Before proceeding with event-based video frame interpolation, it is crucial to temporally align and spatially register both cameras. Hereafter, we describe the components and objectives, and then detail our alignment and registration processes step-by-step.

Let $(x_j, y_j)$ denote a pixel location at time $t_j$ in the event-based camera, and $(x_i, y_i)$ represent the corresponding pixel location at time $t_i$ in the RGB camera. We define the projection function $\mathbf{P}: \mathbb{R}^{H_{\text{event}} \times W_{\text{event}}\times T_{\text{event}}} \rightarrow \mathbb{R}^{H_{\text{image}} \times W_{\text{image}}\times T_{\text{image}}}$ as
\begin{equation}
(x_i, y_i, t_i) = \mathbf{P}(x_j, y_j, t_j)\comma
\end{equation}
where $H_{\text{event}}$, $W_{\text{event}}$, $T_{\text{event}}$, $H_{\text{image}}$, $W_{\text{image}}$, and $T_{\text{image}}$ denote the height, width, and time of the event-based camera image, and the RGB camera image, respectively.
We compute the temporal and spatial projection from the event-based camera to the RGB camera in three steps:

\mysection{1. Temporal synchronization:} 
 We aim to find the event sequence $\mathbf{E}(t_j)$ that aligns temporally with $\mathbf{I}(t_i)$. Since the frame rate of $\mathbf{I}$ is 120 \fps, we first calculate a sequence of accumulated events, denoted by $\mathbf{A()}$, at the same frame rate, and proceed to a visual inspection to align both sequences. This produces a first estimate of $t_i$ and $t_j$. 
 
\begin{figure}[t]
\centering
    \includegraphics[width=\columnwidth]{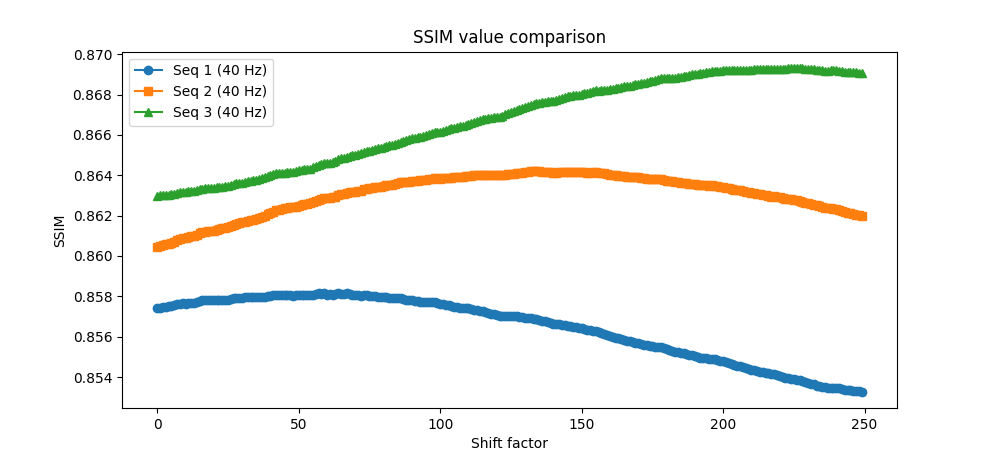}
    \caption{\textbf{SSIM values for different time shifts calculated for three interleaved sequences at a frame rate of 40 \fps.}}
    \label{fig:SSIM}
\end{figure}

 We then reduce the temporal shift $\Delta t_{j\rightarrow i} = t_j-t_i$ as follows. As the greatest common divisor between the sampling rates of $\mathbf{I}()$ and $\mathbf{E}()$ is 40, we split the $\mathbf{I}()$ original sequence into three interleaved sub-sequences $\mathbf{Seq}$ at 40 \fps. Likewise, we generate 250 sequences of accumulated events over a period of 25 ms, starting at $t_j + k\times 100\mu s$ with $k\in \{0,\,...,\,249\}$. Finally, we compute the mean structural similarity index measure (SSIM, see \cite{Wang2004Image}) between the grayscale differences of successive frames of Seq and $\mathbf{A}(t_j + k\times 100\mu s)$ over 10 frames for all values of $k$, and select the sequence and $k$ value that maximize the SSIM. 
 An illustration of the evolution of the SSIM with respect to $k$ is shown in Figure \ref{fig:SSIM}. In that example, we chose Seq 3 and $k$ around 220. This calculation is repeated for each video snapshot, except that we use a previous estimate to skip further visual inspection steps. 
%

\mysection{2. Spatial shift estimation:} Once synchronized temporally, we identify common features between the two images, such as the borders of sports equipment or the athletes. Let $(x_j^{(1)}, y_j^{(1)})$ denote a first feature location in the event-based camera, and $(x_i^{(1)}, y_i^{(1)})$ represent the corresponding feature location in the RGB camera.  We calculate the horizontal and vertical shift as $\Delta_x = x_j^{(1)} - x_i^{(1)}$ and $\Delta_y = y_j^{(1)} - y_i^{(1)}$, respectively.

\begin{figure*}[!t]
    \centering
    \setlength\tabcolsep{0pt} 
    \begin{tabular}{*{8}{c}}
        \adjustbox{valign=m,rotate=90,width=0.5em}{Original 120 \fps} &
        \adjustimage{width=0.133\linewidth,frame=1pt,cfbox=red 1pt 0pt}{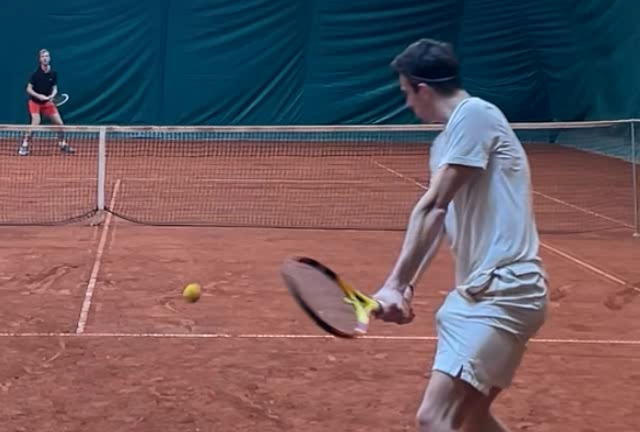} &
        \adjustimage{width=0.133\linewidth,frame=1pt,cfbox=red 1pt 0pt}{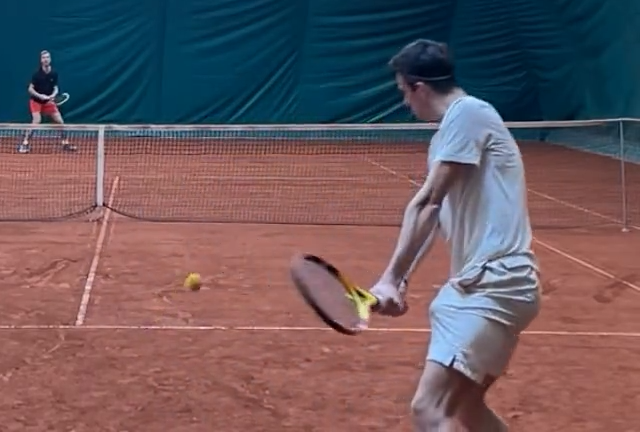} &
        \adjustimage{width=0.133\linewidth,frame=1pt,cfbox=red 1pt 0pt}{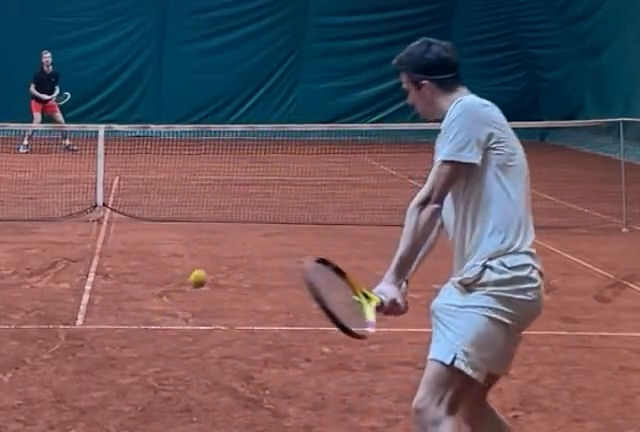} &
        \adjustimage{width=0.133\linewidth,frame=1pt,cfbox=red 1pt 0pt}{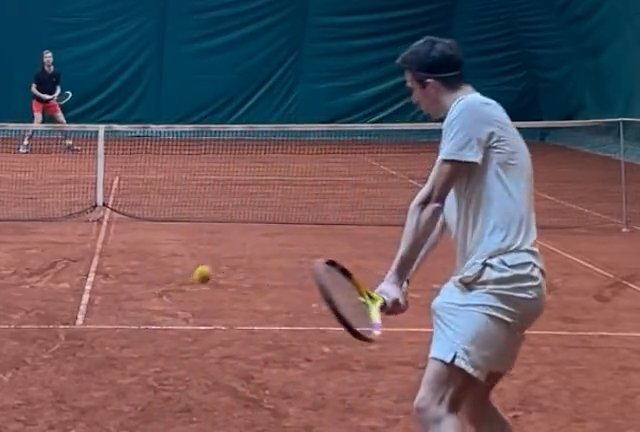} &
        \adjustimage{width=0.133\linewidth,frame=1pt,cfbox=red 1pt 0pt}{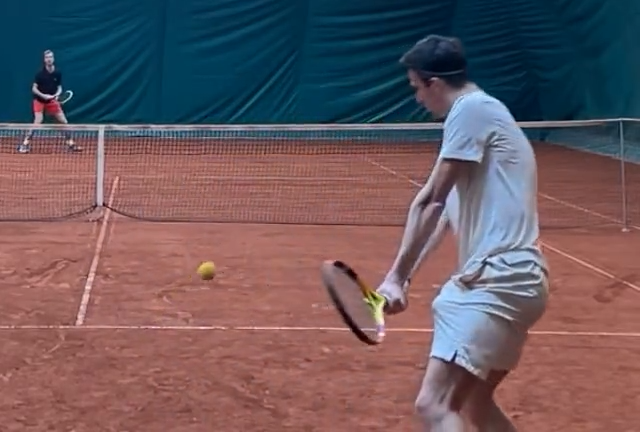} &
        \adjustimage{width=0.133\linewidth,frame=1pt,cfbox=red 1pt 0pt}{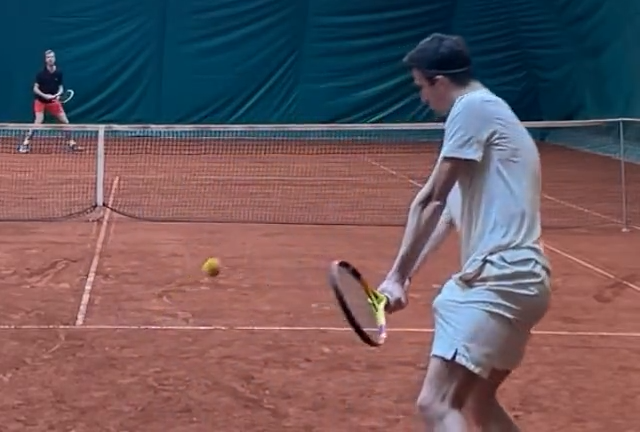} &
        \adjustimage{width=0.133\linewidth,frame=1pt,cfbox=red 1pt 0pt}{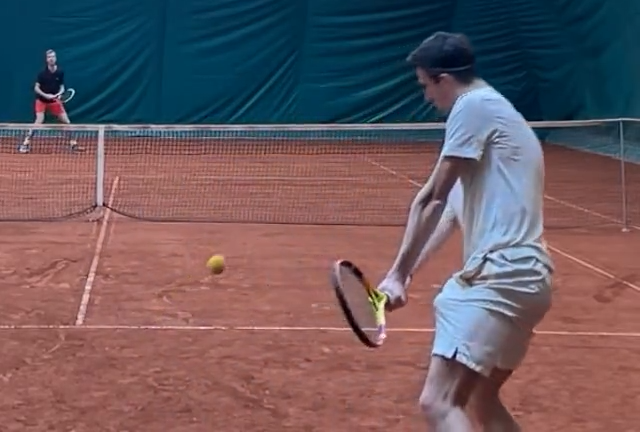} \\

        \adjustbox{valign=m,rotate=90,width=0.5em}{40$\rightarrow$ 120 \fps} &
        \adjustimage{width=0.133\linewidth,frame=1pt,cfbox=red 1pt 0pt}{figure/sequences/sized/120-00.png} &
        \adjustimage{width=0.133\linewidth,frame=1pt,cfbox=green 1pt 0pt}{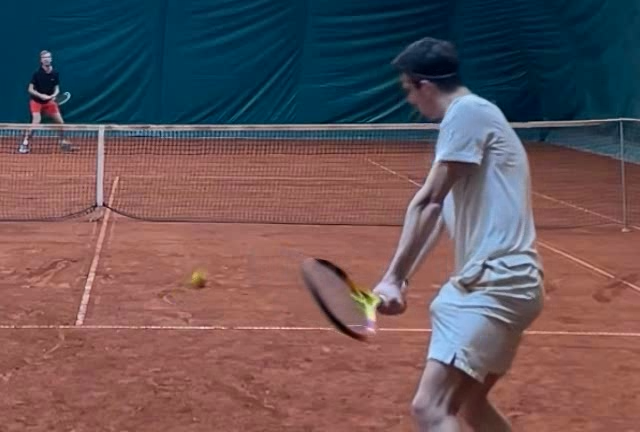} &
        \adjustimage{width=0.133\linewidth,frame=1pt,cfbox=green 1pt 0pt}{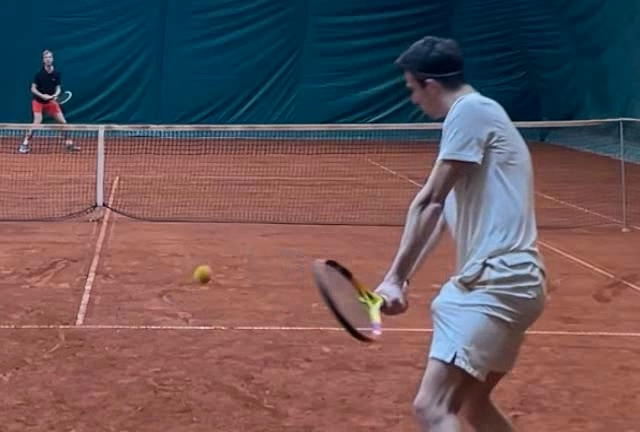} &
        \adjustimage{width=0.133\linewidth,frame=1pt,cfbox=red 1pt 0pt}{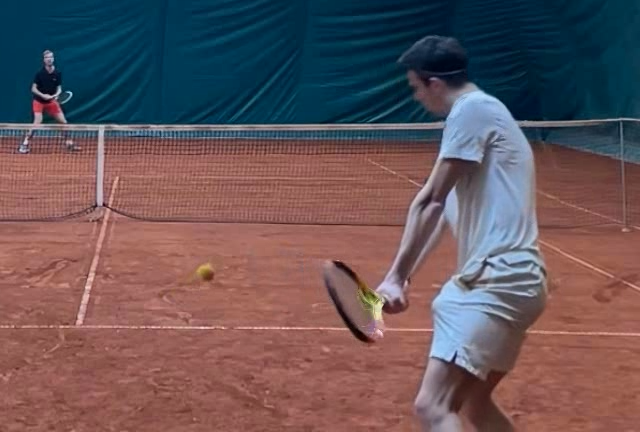} &
        \adjustimage{width=0.133\linewidth,frame=1pt,cfbox=green 1pt 0pt}{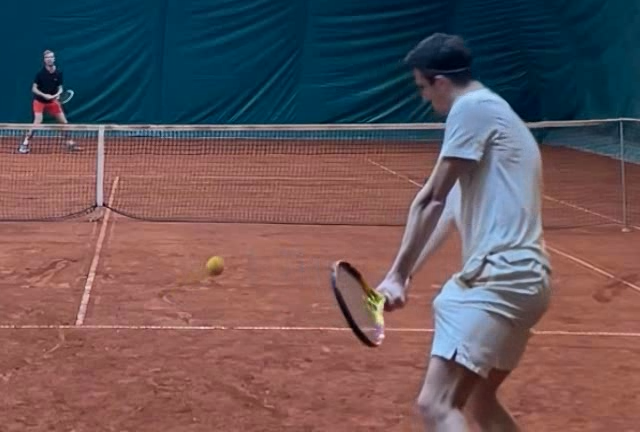} &
        \adjustimage{width=0.133\linewidth,frame=1pt,cfbox=green 1pt 0pt}{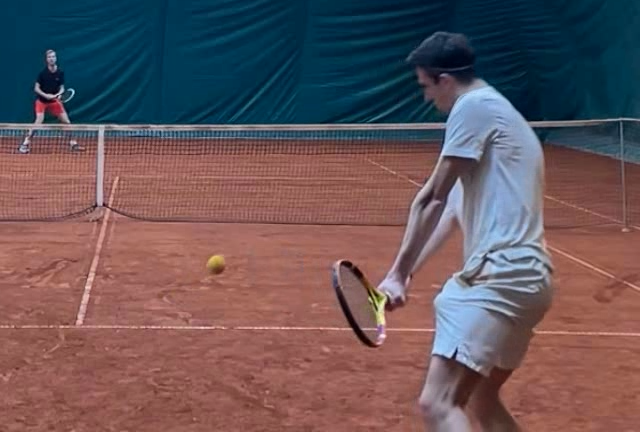} &
        \adjustimage{width=0.133\linewidth,frame=1pt,cfbox=red 1pt 0pt}{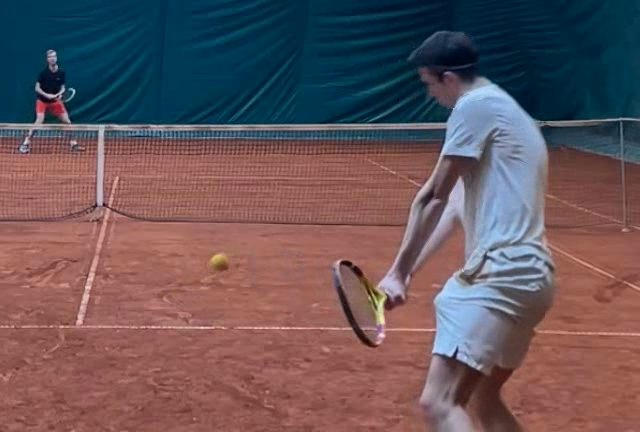} \\

        \adjustbox{valign=m,rotate=90,width=0.5em}{20$\rightarrow$ 120 \fps} &
        \adjustimage{width=0.133\linewidth,frame=1pt,cfbox=red 1pt 0pt}{figure/sequences/sized/120-00.png} &
        \adjustimage{width=0.133\linewidth,frame=1pt,cfbox=green 1pt 0pt}{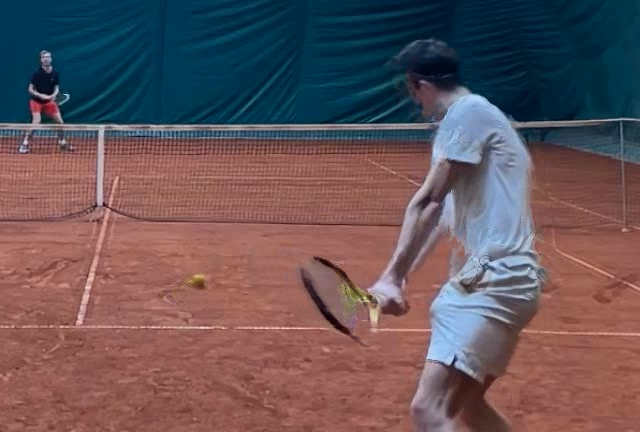} &
        \adjustimage{width=0.133\linewidth,frame=1pt,cfbox=green 1pt 0pt}{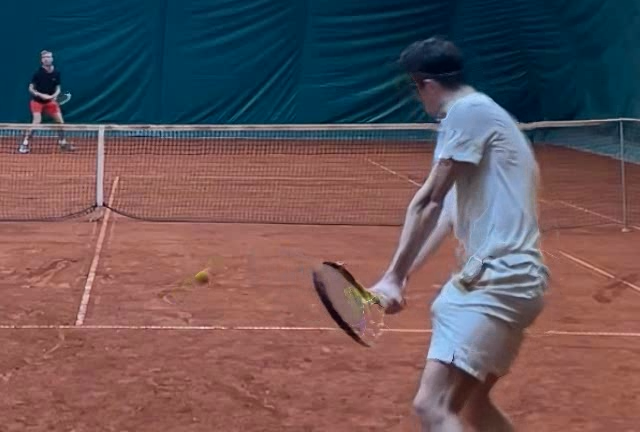} &
        \adjustimage{width=0.133\linewidth,frame=1pt,cfbox=green 1pt 0pt}{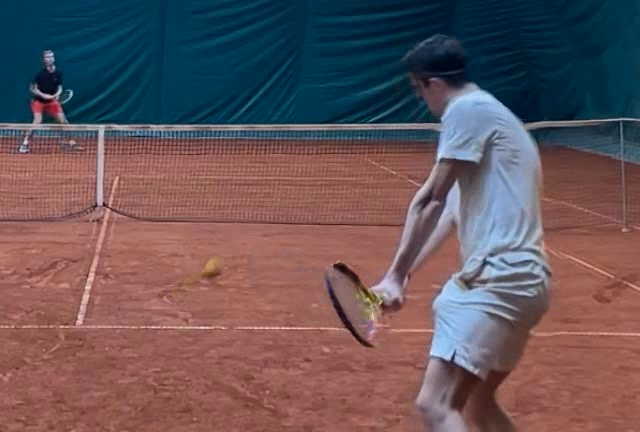} &
        \adjustimage{width=0.133\linewidth,frame=1pt,cfbox=green 1pt 0pt}{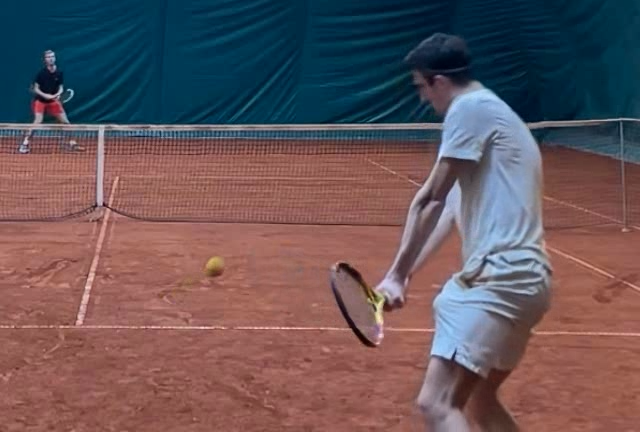} &
        \adjustimage{width=0.133\linewidth,frame=1pt,cfbox=green 1pt 0pt}{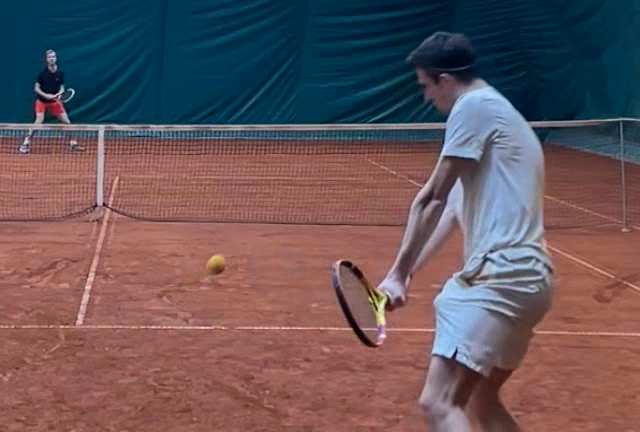} &
        \adjustimage{width=0.133\linewidth,frame=1pt,cfbox=red 1pt 0pt}{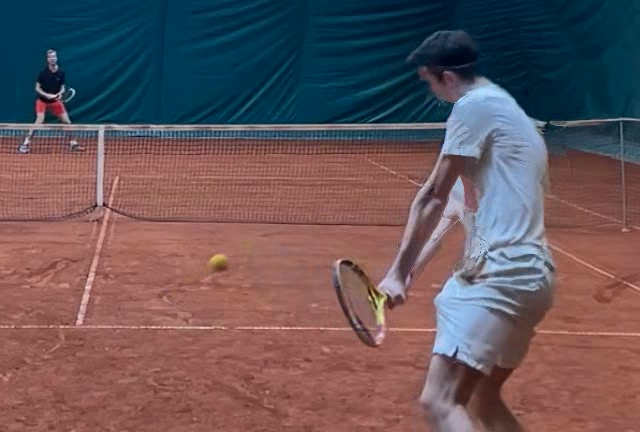} \\

        \adjustbox{valign=m,rotate=90,width=0.5em}{10$\rightarrow$ 120 \fps} &
        \adjustimage{width=0.133\linewidth,frame=1pt,cfbox=red 1pt 0pt}{figure/sequences/sized/120-00.png} &
        \adjustimage{width=0.133\linewidth,frame=1pt,cfbox=green 1pt 0pt}{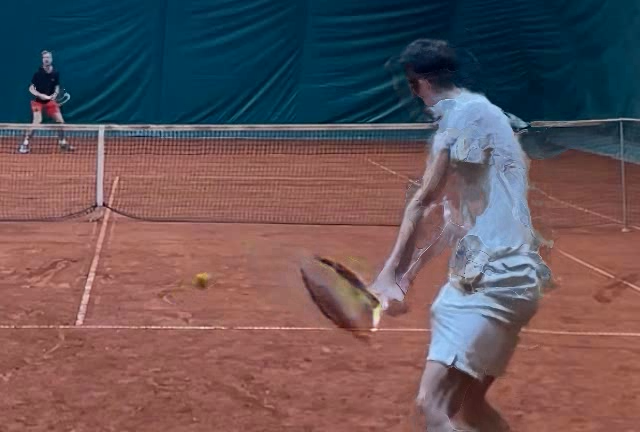} &
        \adjustimage{width=0.133\linewidth,frame=1pt,cfbox=green 1pt 0pt}{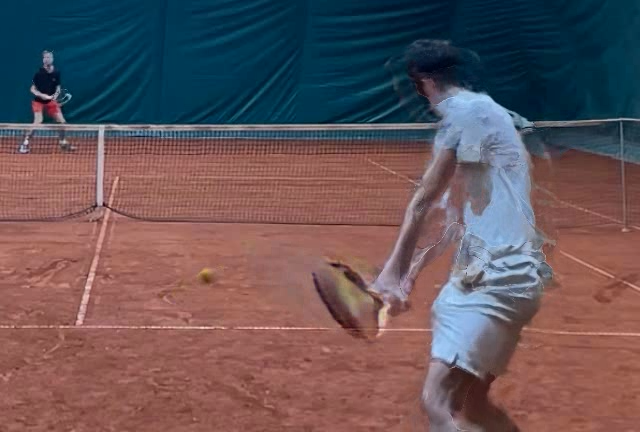} &
        \adjustimage{width=0.133\linewidth,frame=1pt,cfbox=green 1pt 0pt}{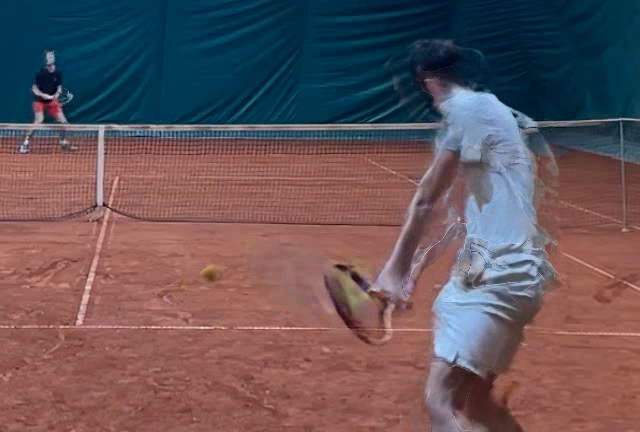} &
        \adjustimage{width=0.133\linewidth,frame=1pt,cfbox=green 1pt 0pt}{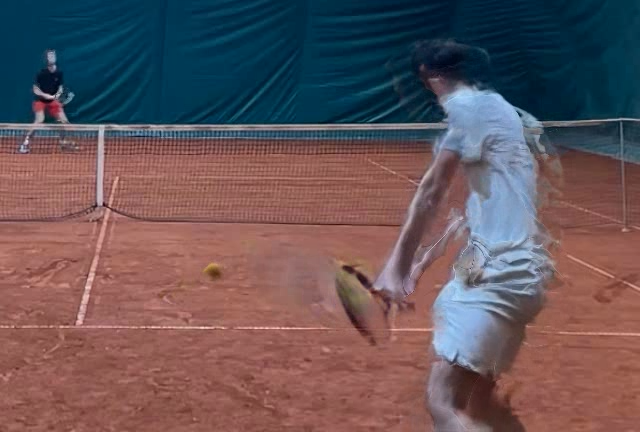} &
        \adjustimage{width=0.133\linewidth,frame=1pt,cfbox=green 1pt 0pt}{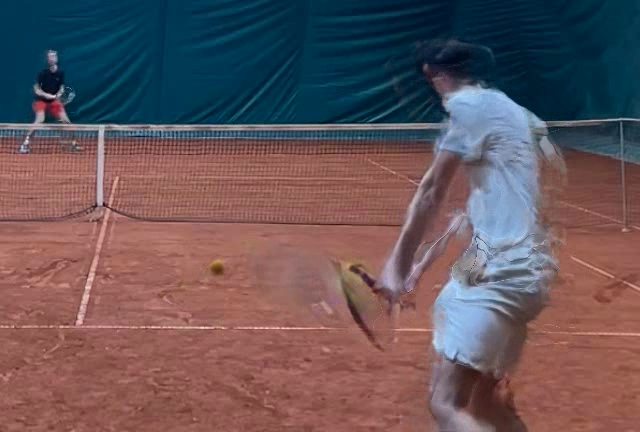} &
        \adjustimage{width=0.133\linewidth,frame=1pt,cfbox=green 1pt 0pt}{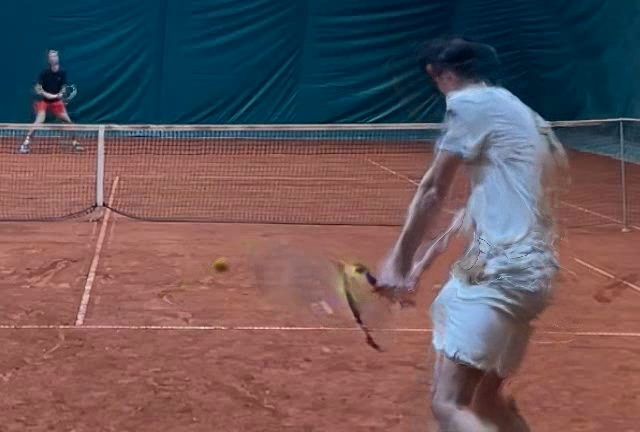} \\
    \end{tabular}
    \caption{\textbf{Qualitative results obtained by Timelens with different input frame rates.} The first line shows the original $120$ \fps video, the second the $40$ \fps with two  intermediate frame interpolated, the third line a $20$ \fps subsampled with $5$ out of $6$ frames interpolated, and finally a $10$ \fps input video with $11$ frames interpolated (only $6$ shown). It can be seen that the tennis ball is well-placed on all interpolated frames. However, for the racquet, the fast movement is only well interpolated with an initial frame rate of $20$ and $40$ while $10$ \fps does not provide satisfactory results. \textcolor{red}{Original} frames are shown in \textcolor{red}{red}, while \textcolor{green}{interpolated} ones are shown in \textcolor{green}{green}. }
    \label{fig:qualitative}
\end{figure*}

\mysection{3. Scale estimation:} Similarly, let $(x_j^{(2)}, x_j^{(2)})$ and $(x_i^{(2)}, y_i^{(2)})$ denote another set of feature locations, we compute the scale ratio between the two images based on the distance between the matched features. Let $\Delta_{\text{event}}$ and $\Delta_{\text{image}}$ denote the Euclidean distances between the matched features in the event-based and RGB cameras, respectively. The scale ratio $r$ is given by
\[
    r = \frac{\Delta_{\text{image}}}{\Delta_{\text{event}}} = \frac{ \sqrt{(x_i^{(2)} - x_i^{(1)})^2 + (y_i^{(2)} - y_i^{(1)})^2 }}{ \sqrt{(x_j^{(2)} - x_j^{(1)})^2 + (y_j^{(2)} - y_j^{(1)})^2 }} \point
    \]
 
The overall projection function then becomes
\begin{equation}
   \mathbf{P} =  
    \begin{cases}
    x_i = r\times(x_j - \Delta_x)\\
    y_j = r\times(y_j - \Delta_y)\\
    t_i = t_j - \Delta_{j\rightarrow i}\comma
    \end{cases}
\end{equation}
which enables the projection of pixel coordinates from the event-based camera to the corresponding locations on the RGB camera, facilitating the generation of intermediate frames for event-based video frame interpolation.

\section{Experiments}
\label{sec:experiments}

\subsection{Experimental setup}
\label{subsec:exprimental_setup}

In this section, we provide a detailed overview of the experimental setup used to generate intermediate frames on our captured data footage to produce slow-motion content. We focus on two main aspects: the initial temporal resolution and upscaling factor, and the summary of the literature method Timelens~\cite{Tulyakov2021TimeLens} that we used in our experiments.


\mysection{Temporal resolution and upscaling.}
To comprehensively evaluate the performance of our method, we explore various combinations of initial temporal resolution and upscaling factors. Given access to a $120$ \fps camera, we conduct experiments with different subsampling rates, including $10$, $20$, $40$, and the original $120$ \fps. Additionally, we test different upscaling factors applied by the event-based VFI method, ranging from $\times3$, $\times6$, $\times10$, $\times12$, and $\times25$. 
By examining a wide range of temporal resolutions and upscaling factors, we aim to assess the robustness and scalability of the event-based VFI method across different input configurations.

\mysection{Timelens.}
Timelens~\cite{Tulyakov2021TimeLens} leverages the complementary strengths of synthesis-based and warping-based interpolation techniques.
The method consists of four modules: warping-based interpolation, warping refinement, interpolation by synthesis, and attention-based averaging. Each module serves a specific purpose, such as estimating optical flow, refining interpolated frames, and blending synthesis-based and warping-based results. The method utilizes a voxel grid representation for encoding event sequences and a backbone architecture based on an hourglass network with skip connections.
Interpolation by synthesis directly regresses new frames given input RGB key frames and event sequences, allowing it to handle changes in lighting. However, it may distort image edges and textures in the presence of noisy or insufficient event data. Warping-based interpolation estimates optical flow from events to warp boundary key frames, making it suitable for handling blur and non-linear motion. The warping refinement module computes refined interpolated frames by estimating residual optical flow between warping-based and synthesis-based results. Finally, the attention averaging module blends synthesis-based and warping-based results to achieve a final interpolation result.
We use the open-source code of Timelens with default parameters for generating our sequences.

\subsection{Qualitative results}
\label{subsec:qualitative}

Qualitative results of the interpolated frames are provided in Figure \ref{fig:qualitative} for all four input frame rates brought up to $120$ \fps. It can be seen that the interpolation process effectively tracks the trajectory of the tennis ball across all interpolated frames, showcasing Timelens' ability in maintaining object consistency. However, the racquet's swift motion demonstrates uneven performance at the initial frame rate. Notably, the $20$ \fps and $40$ \fps inputs exhibit more faithful interpolation of the racquet's trajectory compared to the $10$ \fps input, where evident artifacts emerge. We supplement these observations with additional original frame rates and scale factors, provided in a side-by-side video format for enhanced visualization, in supplementary material~\footnote{Supplementary material: https://bit.ly/3xmhRKU}. Despite the promising outcomes, residual artifacts persist, suggesting avenues for improvement in future iterations of the methodology.

\subsection{Future work}
\label{sec:future}

In future work, we plan to explore several avenues for enhancing the method and extend the scope of our research:

\mysection{Optical Alignment Improvement.} We noticed that misaligned events generate artifacts in the interpolated stream. Therefore, we intend to implement a beam splitter setup to achieve better alignment of the optical centers between the RGB camera and the event-based camera. 

\mysection{Performance enhancement.} We also aim to investigate techniques to further improve the performance of Timelens, particularly when applied to sports-centric videos captured by event-based cameras. This may involve incorporating domain-specific features to enhance the accuracy and realism of the generated slow-motion footage.

\mysection{Exploration of downstream analysis techniques.} Beyond video frame interpolation, we plan to explore downstream analysis techniques enabled by the generated slow-motion footage. One potential avenue is player movement analysis, where we can leverage the detailed temporal information provided by the slow-motion videos to gain insights into player movement and performance.

\section{Conclusion}
\label{sec:conclusion}

In conclusion, this project contributes to the advancement of sports video understanding through the analysis of the potential of event-based video frame interpolation. 
Our primary contributions lie in the recording of sports data, including the alignment of the RGB and event-based camera temporally and spatially. 
Leveraging these recorded sport sequences, we conducted tests of Timelens, an already established event-based video frame interpolation method. Through qualitative analysis, we have provided insights into the performance of Timelens in generating slow-motion footage on our dataset, showcasing its applicability and effectiveness in the context of sports videos captured in dynamic and uncontrolled environments. 
Moving forward, further research in this domain could explore better physical setups, refinements to existing methods with the development of specialized algorithms tailored to the unique characteristics of sports footage, ultimately enhancing the quality and utility of sports video analysis tools.


%
\IEEEpeerreviewmaketitle

\section*{Acknowledgment}
A. Cioppa is funded by the
F.R.S.-FNRS. This work was partly supported by the King Abdullah University of Science and Technology (KAUST) Office of Sponsored Research through the Visual Computing Center (VCC) funding and the SDAIA-KAUST
Center of Excellence in Data Science and Artificial Intelligence (SDAIA-KAUST AI).  We thank Adrien Deli\`ege and Arnaud Leduc, the two semi-pro tennis players, as well as the Royal Tennis Club de Li\`ege for allowing us to record one of their game.



%

\bibliographystyle{ieeetr}
\bibliography{bib/abbreviation-short,
bib/3D, 
bib/action, 
bib/dataset, 
bib/depth, 
bib/depth-related, 
bib/event,
bib/game, 
bib/image,
bib/labo,
bib/learning,
bib/library, 
bib/optical-flow,
bib/soccer,
bib/soccernet-challenge, 
bib/sports, 
bib/vision,
bib/PUT_NEW_REFS_HERE_ONLY}

\begin{thebibliography}{10}

\bibitem{Tulyakov2021TimeLens}
S.~Tulyakov, D.~Gehrig, S.~Georgoulis, J.~Erbach, M.~Gehrig, Y.~Li, and D.~Scaramuzza, ``{Time Lens}: Event-based video frame interpolation,'' in {\em IEEE/CVF Conf. Comput. Vis. Pattern Recognit. (CVPR)}, (Nashville, TN, USA), pp.~16150--16159, Inst. Electr. Electron. Eng. (IEEE), Jun. 2021.

\bibitem{Parihar2021AComprehensive}
A.~S. Parihar, D.~Varshney, K.~Pandya, and A.~Aggarwal, ``A comprehensive survey on video frame interpolation techniques,'' {\em Vis. Comput.}, vol.~38, pp.~295--319, Jan. 2021.

\bibitem{Niklaus2017Video}
S.~Niklaus, L.~Mai, and F.~Liu, ``Video frame interpolation via adaptive separable convolution,'' in {\em IEEE Int. Conf. Comput. Vis. (ICCV)}, (Venice, Italy), pp.~261--270, Inst. Electr. Electron. Eng. (IEEE), Oct. 2017.

\bibitem{Meyer2015Phasebased}
S.~Meyer, O.~Wang, H.~Zimmer, M.~Grosse, and A.~Sorkine-Hornung, ``Phase-based frame interpolation for video,'' in {\em IEEE Conf. Comput. Vis. Pattern Recognit. (CVPR)}, (Boston, MA, USA), pp.~1410--1418, Inst. Electr. Electron. Eng. (IEEE), Jun. 2015.

\bibitem{Niklaus2018ContextAware}
S.~Niklaus and F.~Liu, ``Context-aware synthesis for video frame interpolation,'' in {\em IEEE/CVF Conf. Comput. Vis. Pattern Recognit. (CVPR)}, (Salt Lake City, UT, USA), pp.~1701--1710, Inst. Electr. Electron. Eng. (IEEE), Jun. 2018.

\bibitem{vanAmersfoort2017Frame-arxiv}
J.~van Amersfoort, W.~Shi, A.~Acosta, F.~Massa, J.~Totz, Z.~Wang, and J.~Caballero, ``Frame interpolation with multi-scale deep loss functions and generative adversarial networks,'' {\em arXiv}, vol.~abs/1711.06045, 2017.

\bibitem{EVS2024XtraMotion}
{EVS Broadcast Equipment}, ``{XtraMotion}.'' \url{https://evs.com/products/live-replays-storytelling/xtramotion}.

\bibitem{Xue2019Video}
T.~Xue, B.~Chen, J.~Wu, D.~Wei, and W.~T. Freeman, ``Video enhancement with task-oriented flow,'' {\em Int. J. Comput. Vis.}, vol.~127, pp.~1106--1125, Feb. 2019.

\bibitem{Soomro2012UCF101-arxiv}
K.~Soomro, A.~R. Zamir, and M.~Shah, ``{UCF101}: A dataset of 101 human actions classes from videos in the wild,'' {\em arXiv}, vol.~abs/1212.0402, 2012.

\bibitem{Sim2021XVFI}
H.~Sim, J.~Oh, and M.~Kim, ``{XVFI}: {eXtreme} video frame interpolation,'' in {\em IEEE/CVF Int. Conf. Comput. Vis. (ICCV)}, (Montr{\'e}al, Can.), pp.~14469--14478, Inst. Electr. Electron. Eng. (IEEE), Oct. 2021.

\bibitem{Baker2007ADatabase}
S.~Baker, S.~Roth, D.~Scharstein, M.~J. Black, J.~Lewis, and R.~Szeliski, ``A database and evaluation methodology for optical flow,'' in {\em IEEE Int. Conf. Comput. Vis. (ICCV)}, (Rio de Janeiro, Brazil), pp.~1--8, Inst. Electr. Electron. Eng. (IEEE), 2007.

\bibitem{Choi2020Channel}
M.~Choi, H.~Kim, B.~Han, N.~Xu, and K.~M. Lee, ``Channel attention is all you need for video frame interpolation,'' in {\em AAAI Conf. Artif. Intell.}, vol.~34, pp.~10663--10671, Association for the Advancement of Artificial Intelligence (AAAI), Apr. 2020.

\bibitem{Chen2023SportsSloMo-arxiv}
J.~Chen and H.~Jiang, ``{SportsSloMo}: A new benchmark and baselines for human-centric video frame interpolation,'' {\em arXiv}, vol.~abs/2308.16876, 2023.

\bibitem{Zheng2023Deep-arxiv}
X.~Zheng, Y.~Liu, Y.~Lu, T.~Hua, T.~Pan, W.~Zhang, D.~Tao, and L.~Wang, ``Deep learning for event-based vision: A comprehensive survey and benchmarks,'' {\em arXiv}, vol.~abs/2302.08890, 2023.

\bibitem{Tulyakov2022TimeLens++}
S.~Tulyakov, A.~Bochicchio, D.~Gehrig, S.~Georgoulis, Y.~Li, and D.~Scaramuzza, ``{Time Lens++}: Event-based frame interpolation with parametric nonlinear flow and multi-scale fusion,'' in {\em IEEE/CVF Conf. Comput. Vis. Pattern Recognit. (CVPR)}, (New Orleans, LA, USA), pp.~17734--17743, Inst. Electr. Electron. Eng. (IEEE), Jun. 2022.

\bibitem{Gao2023SuperFast}
Y.~Gao, S.~Li, Y.~Li, Y.~Guo, and Q.~Dai, ``{SuperFast}: 200$\times$ video frame interpolation via event camera,'' {\em IEEE Trans. Pattern Anal. Mach. Intell.}, vol.~45, pp.~7764--7780, Jun. 2023.

\bibitem{Moeslund2014Computer}
T.~B. Moeslund, G.~Thomas, and A.~Hilton, {\em Computer vision in sports}.
\newblock Springer, 2014.

\bibitem{Naik2022AComprehensive}
B.~T. Naik, M.~F. Hashmi, N.~D. Bokde, and Z.~M. Yaseen, ``A comprehensive review of computer vision in sports: Open issues, future trends and research directions,'' {\em Appl. Sci.}, vol.~12, pp.~1--49, Apr. 2022.

\bibitem{Ingwersen2023SportsPose}
C.~K. Ingwersen, C.~M{\o}ller~Mikkelstrup, J.~N. Jensen, M.~Rieger~Hannemose, and A.~B. Dahl, ``{SportsPose} - a dynamic {3D} sports pose dataset,'' in {\em IEEE/CVF Conf. Comput. Vis. Pattern Recognit. Work. (CVPRW)}, (Vancouver, Can.), pp.~5219--5228, Inst. Electr. Electron. Eng. (IEEE), Jun. 2023.

\bibitem{Scott2022SoccerTrack}
A.~Scott, I.~Uchida, M.~Onishi, Y.~Kameda, K.~Fukui, and K.~Fujii, ``{SoccerTrack}: A dataset and tracking algorithm for soccer with fish-eye and drone videos,'' in {\em IEEE/CVF Conf. Comput. Vis. Pattern Recognit. Work. (CVPRW)}, (New Orleans, LA, USA), pp.~3568--3578, Inst. Electr. Electron. Eng. (IEEE), Jun. 2022.

\bibitem{VanZandycke2022DeepSportradarv1}
G.~Van~Zandycke, V.~Somers, M.~Istasse, C.~D. Don, and D.~Zambrano, ``{DeepSportradar}-v1: Computer vision dataset for sports understanding with high quality annotations,'' in {\em Int. ACM Work. Multimedia Content Anal. Sports (MMSports)}, (Lisbon, Port.), pp.~1--8, ACM, Oct. 2022.

\bibitem{Istasse2023DeepSportradarv2}
M.~Istasse, V.~Somers, P.~Elancheliyan, J.~De, and D.~Zambrano, ``{DeepSportradar}-v2: A multi-sport computer vision dataset for sport understandings,'' in {\em Int. ACM Work. Multimedia Content Anal. Sports (MMSports)}, (Ottawa, Ontario, Can.), pp.~23--29, ACM, Oct. 2023.

\bibitem{Giancola2022SoccerNet}
S.~Giancola, A.~Cioppa, A.~Deli{\`e}ge, F.~Magera, V.~Somers, L.~Kang, X.~Zhou, O.~Barnich, C.~De~Vleeschouwer, A.~Alahi, B.~Ghanem, M.~Van~Droogenbroeck, and \etal, ``{SoccerNet} 2022 challenges results,'' in {\em Int. ACM Work. Multimedia Content Anal. Sports (MMSports)}, (Lisbon, Port.), pp.~75--86, ACM, Oct. 2022.

\bibitem{Cioppa2023SoccerNetChallenge-arxiv}
A.~Cioppa, S.~Giancola, V.~Somers, F.~Magera, X.~Zhou, H.~Mkhallati, A.~Deli{\`e}ge, J.~Held, C.~Hinojosa, A.~M. Mansourian, P.~Miralles, O.~Barnich, C.~De~Vleeschouwer, A.~Alahi, B.~Ghanem, M.~Van~Droogenbroeck, and \etal, ``{SoccerNet} 2023 challenges results,'' {\em arXiv}, vol.~abs/2309.06006, 2023.

\bibitem{Giancola2018SoccerNet}
S.~Giancola, M.~Amine, T.~Dghaily, and B.~Ghanem, ``{SoccerNet}: A scalable dataset for action spotting in soccer videos,'' in {\em IEEE/CVF Conf. Comput. Vis. Pattern Recognit. Work. (CVPRW)}, (Salt Lake City, UT, USA), pp.~1792--179210, Inst. Electr. Electron. Eng. (IEEE), Jun. 2018.

\bibitem{Deliege2021SoccerNetv2}
A.~Deli{\`e}ge, A.~Cioppa, S.~Giancola, M.~J. Seikavandi, J.~V. Dueholm, K.~Nasrollahi, B.~Ghanem, T.~B. Moeslund, and M.~Van~Droogenbroeck, ``{SoccerNet}-v2: A dataset and benchmarks for holistic understanding of broadcast soccer videos,'' in {\em IEEE Int. Conf. Comput. Vis. Pattern Recognit. Work. (CVPRW), CVsports}, (Nashville, TN, USA), pp.~4508--4519, Jun. 2021.

\bibitem{Cioppa2022Scaling}
A.~Cioppa, A.~Deli{\`e}ge, S.~Giancola, B.~Ghanem, and M.~Van~Droogenbroeck, ``Scaling up {SoccerNet} with multi-view spatial localization and re-identification,'' {\em Sci. Data}, vol.~9, pp.~1--9, Jun. 2022.

\bibitem{Cioppa2022SoccerNetTracking}
A.~Cioppa, S.~Giancola, A.~Deliege, L.~Kang, X.~Zhou, Z.~Cheng, B.~Ghanem, and M.~Van~Droogenbroeck, ``{SoccerNet}-tracking: Multiple object tracking dataset and benchmark in soccer videos,'' in {\em IEEE Int. Conf. Comput. Vis. Pattern Recognit. Work. (CVPRW), CVsports}, (New Orleans, LA, USA), pp.~3490--3501, Inst. Electr. Electron. Eng. (IEEE), Jun. 2022.

\bibitem{Held2023VARS}
J.~Held, A.~Cioppa, S.~Giancola, A.~Hamdi, B.~Ghanem, and M.~Van~Droogenbroeck, ``{VARS}: Video assistant referee system for automated soccer decision making from multiple views,'' in {\em IEEE/CVF Conf. Comput. Vis. Pattern Recognit. Work. (CVPRW)}, (Vancouver, Can.), pp.~5086--5097, Inst. Electr. Electron. Eng. (IEEE), Jun. 2023.

\bibitem{Mkhallati2023SoccerNetCaption}
H.~Mkhallati, A.~Cioppa, S.~Giancola, B.~Ghanem, and M.~Van~Droogenbroeck, ``{SoccerNet}-caption: Dense video captioning for soccer broadcasts commentaries,'' in {\em IEEE/CVF Conf. Comput. Vis. Pattern Recognit. Work. (CVPRW)}, (Vancouver, Can.), pp.~5074--5085, Inst. Electr. Electron. Eng. (IEEE), Jun. 2023.

\bibitem{Vandeghen2022SemiSupervised}
R.~Vandeghen, A.~Cioppa, and M.~Van~Droogenbroeck, ``Semi-supervised training to improve player and ball detection in soccer,'' in {\em IEEE Int. Conf. Comput. Vis. Pattern Recognit. Work. (CVPRW), CVsports}, (New Orleans, LA, USA), pp.~3480--3489, Inst. Electr. Electron. Eng. (IEEE), Jun. 2022.

\bibitem{Seweryn2024Improving-arxiv}
K.~Seweryn, G.~Che{\' c}, S.~{\L}ukasik, and A.~Wr{\'o}blewska, ``Improving object detection quality in football through super-resolution techniques,'' {\em arXiv}, vol.~abs/2402.00163, 2024.

\bibitem{Boeker2023Soccer}
M.~Boeker and C.~Midoglu, ``Soccer athlete data visualization and analysis with an interactive dashboard,'' in {\em Int. Conf. Multimedia Retr.}, vol.~13833 of {\em Lect. Notes Comput. Sci.}, pp.~565--576, Springer Int. Publ., 2023.

\bibitem{Cioppa2019ARTHuS}
A.~Cioppa, A.~Deliege, M.~Istasse, C.~De~Vleeschouwer, and M.~Van~Droogenbroeck, ``{ARTHuS}: Adaptive real-time human segmentation in sports through online distillation,'' in {\em IEEE Int. Conf. Comput. Vis. Pattern Recognit. Work. (CVPRW), CVsports}, (Long Beach, CA, USA), pp.~2505--2514, Inst. Electr. Electron. Eng. (IEEE), Jun. 2019.

\bibitem{Maglo2022Efficient}
A.~Maglo, A.~Orcesi, and Q.-C. Pham, ``Efficient tracking of team sport players with few game-specific annotations,'' in {\em IEEE/CVF Conf. Comput. Vis. Pattern Recognit. Work. (CVPRW)}, (New Orleans, LA, USA), pp.~3460--3470, Inst. Electr. Electron. Eng. (IEEE), Jun. 2022.

\bibitem{ArbuesSanguesa2020Using}
A.~Arbu{\'e}s~Sang{\"u}esa, A.~Mart{\'i}n, J.~Fern{\'a}ndez, C.~Ballester, and G.~Haro, ``Using player's body-orientation to model pass feasibility in soccer,'' in {\em IEEE/CVF Conf. Comput. Vis. Pattern Recognit. Work. (CVPRW)}, (Seattle, WA, USA), pp.~3875--3884, Inst. Electr. Electron. Eng. (IEEE), Jun. 2020.

\bibitem{Cioppa2018ABottomUp}
A.~Cioppa, A.~Deli\`ege, and M.~Van~Droogenbroeck, ``A bottom-up approach based on semantics for the interpretation of the main camera stream in soccer games,'' in {\em IEEE Int. Conf. Comput. Vis. Pattern Recognit. Work. (CVPRW), CVsports}, (Salt Lake City, UT, USA), pp.~1846--1855, Jun. 2018.

\bibitem{Soares2022Temporally}
J.~V.~B. Soares, A.~Shah, and T.~Biswas, ``Temporally precise action spotting in soccer videos using dense detection anchors,'' in {\em IEEE Int. Conf. Image Process. (ICIP)}, (Bordeaux, France), pp.~2796--2800, Inst. Electr. Electron. Eng. (IEEE), Oct. 2022.

\bibitem{Hong2022Spotting-arxiv}
J.~Hong, H.~Zhang, M.~Gharbi, M.~Fisher, and K.~Fatahalian, ``Spotting temporally precise, fine-grained events in video,'' {\em arXiv}, vol.~abs/2207.10213, 2022.

\bibitem{Wu2020Fusing-arxiv}
L.~Wu, Z.~Yang, Q.~Wang, M.~Jian, B.~Zhao, J.~Yan, and C.~W. Chen, ``Fusing motion patterns and key visual information for semantic event recognition in basketball videos,'' {\em arXiv}, vol.~abs/2007.06288, 2020.

\bibitem{Denize2023COMEDIAN-arxiv}
J.~Denize, M.~Liashuha, J.~Rabarisoa, A.~Orcesi, and R.~H{\'e}rault, ``{COMEDIAN}: Self-supervised learning and knowledge distillation for action spotting using transformers,'' {\em arXiv}, vol.~abs/2309.01270, 2023.

\bibitem{Seweryn2023Survey-arxiv}
K.~Seweryn, A.~Wr{\'o}blewska, and S.~{\L}ukasik, ``Survey of action recognition, spotting and spatio-temporal localization in soccer -- current trends and research perspectives,'' {\em arXiv}, vol.~abs/2309.12067, 2023.

\bibitem{Cioppa2021Camera}
A.~Cioppa, A.~Deli{\`e}ge, S.~Giancola, F.~Magera, O.~Barnich, B.~Ghanem, and M.~Van~Droogenbroeck, ``Camera calibration and player localization in {SoccerNet-v2} and investigation of their representations for action spotting,'' in {\em IEEE Int. Conf. Comput. Vis. Pattern Recognit. Work. (CVPRW), CVsports}, (Nashville, TN, USA), pp.~4532--4541, Jun. 2021.

\bibitem{Cioppa2020AContextaware}
A.~Cioppa, A.~Deli\`ege, S.~Giancola, B.~Ghanem, M.~Van~Droogenbroeck, R.~Gade, and T.~B. Moeslund, ``A context-aware loss function for action spotting in soccer videos,'' in {\em IEEE/CVF Conf. Comput. Vis. Pattern Recognit. (CVPR)}, (Seattle, WA, USA), pp.~13123--13133, Inst. Electr. Electron. Eng. (IEEE), Jun. 2020.

\bibitem{Gautam2022Assisting-RG}
S.~Gautam, C.~Midoglu, S.~S. Sabet, D.~B.~K. Kshatri, and P.~Halvorsen, ``Assisting soccer game summarization via audio intensity analysis of game highlights,'' in {\em IOE Graduate Conference}, vol.~12, pp.~25--32, Oct. 2022.

\bibitem{Midoglu2022MMSys}
C.~Midoglu, S.~Hicks, V.~Thambawita, T.~Kupka, and P.~Halvorsen, ``{MMSys'22} grand challenge on {AI}-based video production for soccer,'' in {\em ACM Multimedia Systems Conference (MMSys)}, (Athlone, Ireland), pp.~1--6, Jun. 2022.

\bibitem{Sarkhoosh2024AIBased}
M.~H. Sarkhoosh, S.~M.~M. Dorcheh, C.~Midoglu, S.~S. Sabet, T.~Kupka, D.~Johansen, M.~A. Riegler, and P.~Halvorsen, ``{AI}-based cropping of soccer videos for different social media representations,'' in {\em Int. Conf. Multimedia Retr.}, vol.~14557 of {\em Lect. Notes Comput. Sci.}, pp.~279--287, Springer Nat. Switz., 2024.

\bibitem{Wang2004Image}
Z.~Wang, A.~C. Bovik, H.~R. Sheikh, and E.~P. Simoncelli, ``Image quality assessment: From error visibility to structural similarity,'' {\em IEEE Trans. Image Process.}, vol.~13, pp.~600--612, Apr. 2004.

\end{thebibliography}

\end{document}